\documentclass[letterpaper, 10 pt, conference]{ieeeconf}  

\IEEEoverridecommandlockouts                              

\makeatletter
\let\NAT@parse\undefined
\makeatother
\usepackage[numbers,sort&compress]{natbib}
\usepackage[pdftex]{graphicx}
\usepackage{amsmath} 
\usepackage{amssymb}  
\usepackage{multirow}
\usepackage{array,booktabs}
\usepackage{diagbox}

\usepackage{caption}
\usepackage{subcaption}
\usepackage[ruled,linesnumbered]{algorithm2e}
\SetKwInput{KwInput}{Input}
\usepackage{textcomp}
\usepackage{rpm_SIunits}
\usepackage{rpm_acronyms}
\usepackage{rpm_math}
\usepackage{rpm_misc}
\usepackage[font=scriptsize]{caption}
\usepackage{soul}
\usepackage{color}

\usepackage{xcolor}

\newcommand{\gr}[1]{{\textcolor{black}{#1}}}
\newcommand{\bu}[1]{{\textcolor{black}{#1}}}
\usepackage{lipsum}

\title{\LARGE \bf
Edge-guided Multi-domain RGB-to-TIR image Translation

for Training Vision Tasks with Challenging Labels  
}

\author{Dong-Guw Lee$^{1}$, Myung-Hwan Jeon$^{2}$, Younggun Cho$^{3}$, Ayoung Kim$^{1*}$
\thanks{$^\dagger$This research was jointly supported by Korea Institute for Advancement of Technology(KIAT) grant funded by the Korea Government(MOTIE) (P0020536, HRD Program for Industrial Innovation) and Institute of Information \& communications Technology Planning \& Evaluation (IITP) grant funded by the Korea government(MSIT) (No.2022-0-00480, Development of Training and Inference Methods for Goal-Oriented Artificial Intelligence Agents) and (No.2022-0-00448, Deep Total Recall).} 
\thanks{$^{1}$ D. Lee and A. Kim are with the Dept. of Mechanical Engineering, SNU, Seoul, S. Korea {\tt\footnotesize [donkeymouse, ayoungk]@snu.ac.kr}}%
\thanks{$^{2}$ M. Jeon is with the Institute of Advanced Machines and Design, SNU, Seoul, S.Korea {\tt\footnotesize myunghwan.jeon@snu.ac.kr}}%
\thanks{$^{3}$ Y. Cho is with the Dept. Electrical Engineering of Inha University, Incheon, S. Korea {\tt\footnotesize yg.cho@inha.ac.kr}}%
}

\begin{document}

\maketitle
\thispagestyle{empty}
\pagestyle{empty}

\begin{abstract}
\bu{The insufficient number of annotated thermal infrared (TIR) image datasets not only hinders TIR image-based deep learning networks to have comparable performances to that of RGB but it also limits the supervised learning of TIR image-based tasks with challenging labels. As a remedy, we propose a modified multidomain RGB to TIR image translation model focused on edge preservation to employ annotated RGB images with challenging labels. Our proposed method not only preserves key details in the original image but also leverages the optimal TIR style code to portray accurate TIR characteristics in the translated image, when applied on both synthetic and real world RGB images. Using our translation model, we have enabled the supervised learning of deep TIR image-based optical flow estimation and object detection that ameliorated in deep TIR optical flow estimation by reduction in end point error by 56.5\% on average and the best object detection mAP of 23.9\% respectively. Our code and supplementary materials are available at https://github.com/rpmsnu/sRGB-TIR.
}
\end{abstract}

\section{INTRODUCTION} \label{sec:intro}

Recent works on robotics and computer vision have raised interests in the use of \ac{TIR} imaging due to its robustness in environments with harsh weather and poor illumination. Despite this perceptual robustness, \ac{TIR} cameras induce images with low contrast, poor resolution, and ambiguous object boundaries. In particular, such characteristics instigate performance degradation when applying traditional computer vision methods to \ac{TIR} images \cite{tir_calibration}. Deep learning is used as an alternative option to overcome such limitation, yet existing models trained on RGB images hardly adapt well to \ac{TIR} images, and the number of annotated \ac{TIR} image datasets are insufficient for training \ac{TIR}-image-based models with sufficient performance for various tasks. 

\gr{Different solutions are available to overcome lack of well-annotated labels for \ac{TIR}. One way is to employ manual annotations by humans, but it is expensive and labor-intensive. Another way is to obtain annotated data from computer simulations, however, obtaining realistic \ac{TIR} data require sophisticated \ac{TIR} object priors \cite{ganisbetter}. Recent method addresses the problem with self-supervised methods \cite{thermalraft}, while careful initialization and pseudo-labeling were required for comparable performances. }

\begin{figure}[t] 
  \begin{center}
    \includegraphics[width=\linewidth]{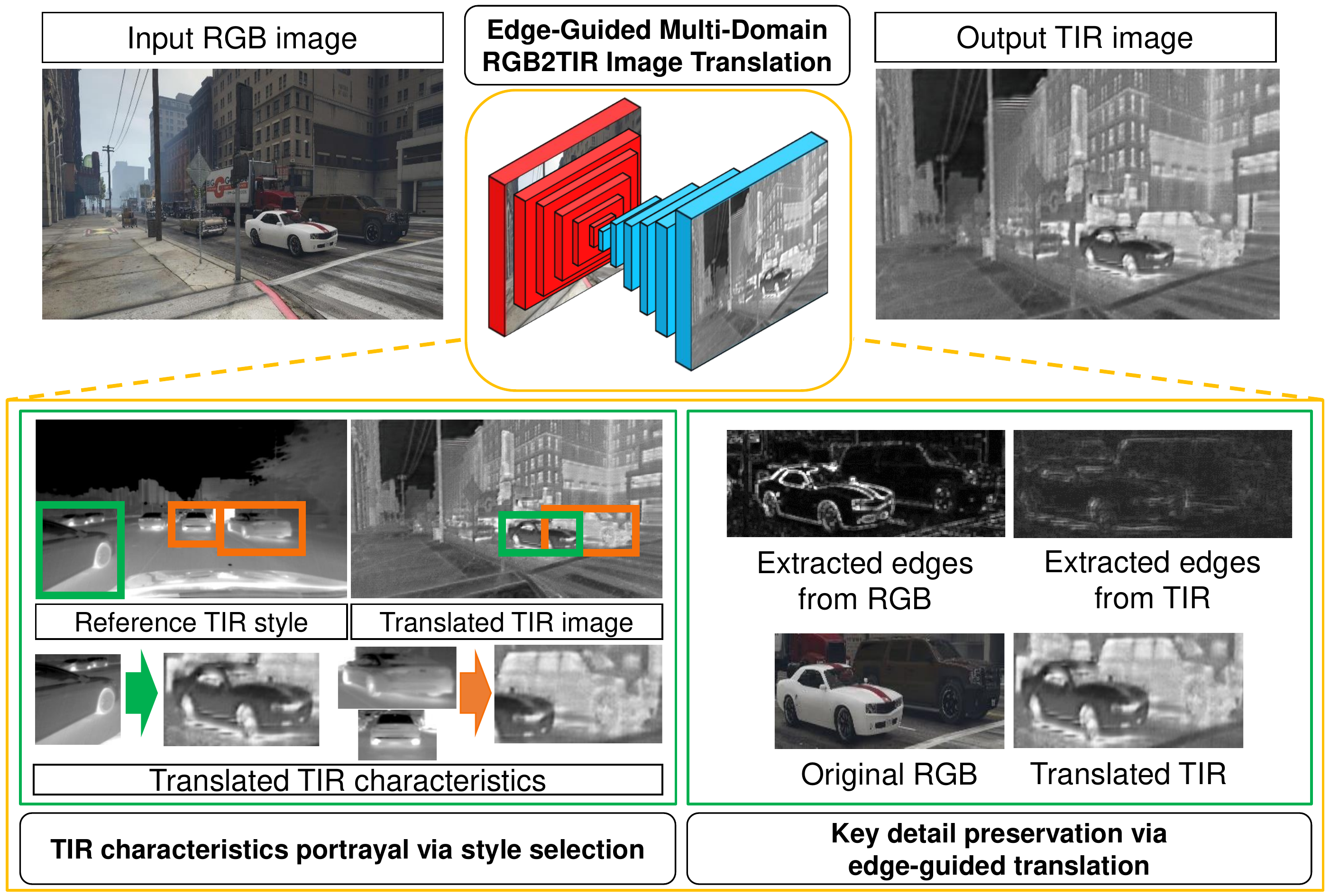}
  \end{center}
  \caption{Our proposed RGB to TIR translation network not only preserves key details in the translated images via edge-guided translation but it also portrays characteristics of thermal images, \l{e.g. the undercarriage of the car is heated in the translated image just like the original TIR image.}}
  \label{overview}
\end{figure}

\gr{Alternative solutions are proposed in \cite{devaguptapu2019borrow,tracking,thermalsegmentationsynthetic} who utilized \ac{GAN}-based image translation methods; these methods obtain annotated \ac{TIR} image data from translating RGB images into \ac{TIR} images and leveraging annotations from RGB data. In fact, \citeauthor{ganisbetter} \cite{ganisbetter} argues that for \ac{TIR} image-based semantic segmentation, leveraging GAN-based translation is much simpler but more accurate way to account for the lack of data than obtaining synthetic \ac{TIR} data from simulations.}


\bu{
GAN-based RGB to \ac{TIR} translation methods have actively been studied in the past \cite{luo2022clawgan,nyberg2018unpaired,ozkanouglu2022infragan,thermalgan,zhang2021wggan}, yet existing \textit{bi-domain}-based image translation methods leverage deterministic bijective mapping function to directly map each pixel in the RGB domain to that of \ac{TIR}. As a consequence, such methods can incur artifacts in the translated \ac{TIR} images when input images differ greatly from the training images and are not suitable translation method for tasks that require image style consistency between two input frames. Contrastive learning-based methods \cite{park2020cut,han2021dual} tried to overcome these shortcomings, yet matching semantic relations between two domains with large semantic discrepancies can be difficult without additional constraints or loss functions. More importantly, translated image styles cannot be controlled with such bi-domain-based methods.    
}

\bu{
On the other hand, we propose an edge-guided \textit{multi-domain} image translation network to translate RGB to \ac{TIR} images. Multi-domain translation network refers to methods that use disentangled content and style latent vectors for image translation \cite{multidomain_definition}. Consequently, artifacts in the translated images could be minimized by selecting the suitable \ac{TIR} style for target RGB image, and image consistency between two consecutive image frames can be maintained. In addition, edge-guided loss is used to preserve the key dynamic details in the translated \ac{TIR} image. The two highlights of our proposed method are illustrated in \figref{overview}.
}


%


Most importantly, no existing study has hardly leveraged RGB to \ac{TIR} translation to train to tasks with extreme level of difficulty in manual labeling.
Using our translation network, we not only enabled supervised learning on tasks with challenging labels such as deep optical flow estimation in \ac{TIR}, but inspired by \cite{radarteacher}, we also validated the effectiveness of our proposed method in object detection without the need for any manual annotations.  




Key contributions of our research as follows:

\begin{enumerate}
  \item We propose an edge-guided and style-controlled multi-domain RGB to \ac{TIR} translation network. Unlike previous methods, our proposed method leverages the most suitable style vector to generate realistic \ac{TIR} images with minimum artifacts and also preserve key details in the image regardless of any \ac{TIR} image styles. Objects in the translated \ac{TIR} images displayed valid edge consistency to both synthetic and real RGB images.

  

  \item We complete a supervised deep \ac{TIR} optical flow estimation, which is a task with an extreme level of difficulty in manual labeling, at a much lesser effort through the dataset generation with our proposed method. By attaining reliable geometric consistency of the objects between consecutive sequential frames in the translated \ac{TIR} images, we demonstrated an amelioration in learning-based \ac{TIR} optical flow estimation.

\item We further validate that our translation model can be extended to semantic tasks such as object detection and lessen the human user's effort in annotations. Our RGB to \ac{TIR} translation pipeline will be open-sourced for future academic research related to robotics and computer vision. 
  
  

  

\end{enumerate}
\section{RELATED WORKS} \label{sec:related}

\subsection{Unpaired RGB to \ac{TIR} image translation}

\begin{figure}[] 
  \begin{center}
    \includegraphics[width=\columnwidth]{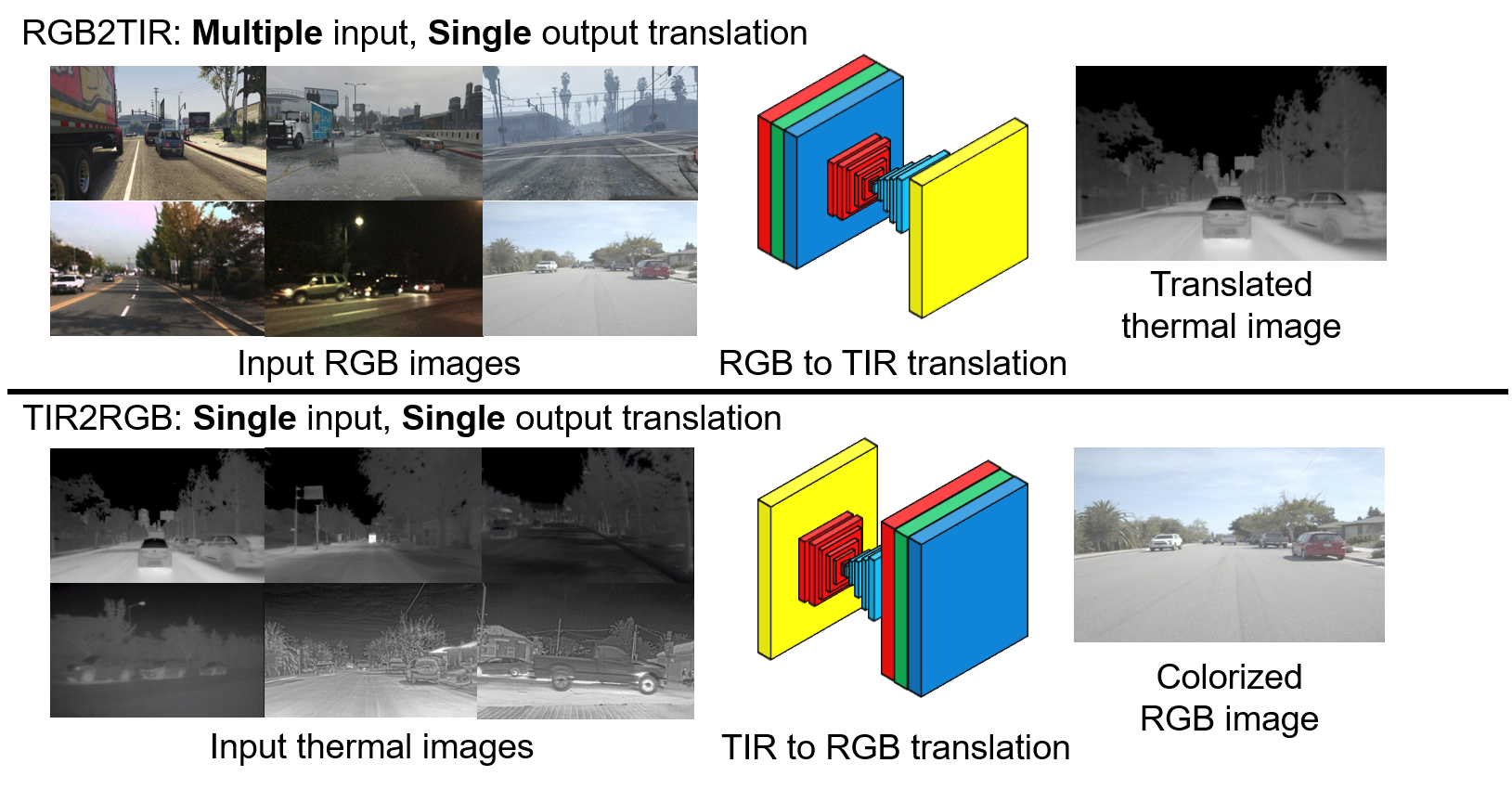}
  \end{center}
  \caption{Main objectives for RGB to \ac{TIR}  and \ac{TIR} to RGB image translation. The former can be defined as multi-input single-output translation problem, and the latter can be viewed as a single input single output translation problem due to the lesser chrominance variation in \ac{TIR} images.}
  \label{problem_definition}
\end{figure}

\gr{Several past studies have addressed image translation from both RGB to \ac{TIR} \cite{ozkanouglu2022infragan,thermalgan,thermalsegmentationsynthetic, tracking, luo2022clawgan,nyberg2018unpaired,zhang2021wggan, rgbtotir_paired} and \ac{TIR} to RGB \cite{devaguptapu2019borrow, tir2rgb1, tir2rgb2, pearlgan}, yet, their objectives are completely different. As illustrated in \figref{problem_definition}, in the former, the translation network needs to cope with several variations that could exist in the input RGB image, but in the generated output, only monochromatic \ac{TIR} images are considered. Hence, the translation can be formulated as a multiple input, single output problem. In the latter, however, diverse chromatic translation of pseudo-RGB images is hardly required, thereby formulating single input, single output problem. Considering such definition, we argue that multi-domain translation methods should be leveraged for RGB to \ac{TIR} image translations.}

\gr{Despite the need for multi-domain-based translation, recent RGB to \ac{TIR} translation were based on bi-domain based translation. As a result, they generalized poorly on images that are outside of the training data, and hardly any key details were preserved in the translated image, especially for those methods based on paired image translation model \cite{thermalsegmentationsynthetic,rgbtotir_paired,ozkanouglu2022infragan,thermalgan}. Especially, these methods were vulnerable to the translation of images with adverse illumination and weather conditions.   }

On the contrary, our edge-guided multi-domain translation method is more robust to translating unseen RGB images and preserves the key details in the translated image. 



\subsection{Deep \ac{TIR} optical flow estimation}

Recent studies on thermal-inertial odometry \cite{thermalraft, trigoni2020deeptio, trigoni2021graph} have actively incorporated deep learning-based \ac{TIR} optical flow estimation models due to their superior performance over the classical methods \cite{survey_of}. However, due to the extreme difficulty in obtaining ground truths flow, even for RGB images \cite{kitti_flow}, deep optical flow estimation model, preinitialized on synthetic RGB images, were used to compute optical flow from \ac{TIR} images \cite{trigoni2020deeptio,trigoni2021graph}. Consequently, erroneous \ac{TIR} optical flow could be expected due to the domain discrepancies between \ac{TIR} and synthetic RGB images.  

On the contrary, we enabled the training of deep \ac{TIR} optical flow estimation model in a supervised manner. Our proposed method not only surpassed the optical flow estimation that adhered to that of \citeauthor{trigoni2020deeptio} \cite{trigoni2020deeptio}, but also no additional separate RGB teacher networks are needed. In addition, compared with self-supervised methods, \bu{our method does not need any explicit and carefully designed pseudo flow generation} \cite{thermalraft}.

\section{Proposed Method}

\subsection{Network overview}

\subsubsection{Key Notations} Listed below are the notation used in the paper.

\begin{itemize}
  \item $G_{TIR}$, $G_{RGB}$, $D_{TIR}$: \ac{TIR} decoder, RGB decoder, \ac{TIR} discriminator. The decoder acts as the generator.
  \item $x_{RGB}$, $x_{TIR}$: Input RGB and \ac{TIR} images.
  \item $E_{RGB}^{c}$, $E_{RGB}^{s}$ and $E_{TIR}^{c}$, $E_{TIR}^{s}$: Content and style encoder for RGB and \ac{TIR} images.
  \item $c_{RGB},s_{RGB}$ and $c_{TIR},s_{TIR}$: Latent content and style vector for RGB and \ac{TIR} images. 
  
\end{itemize}


\subsubsection{Network Architecture}

\begin{figure*}[t] 
  \begin{center}
    \includegraphics[width=\linewidth]{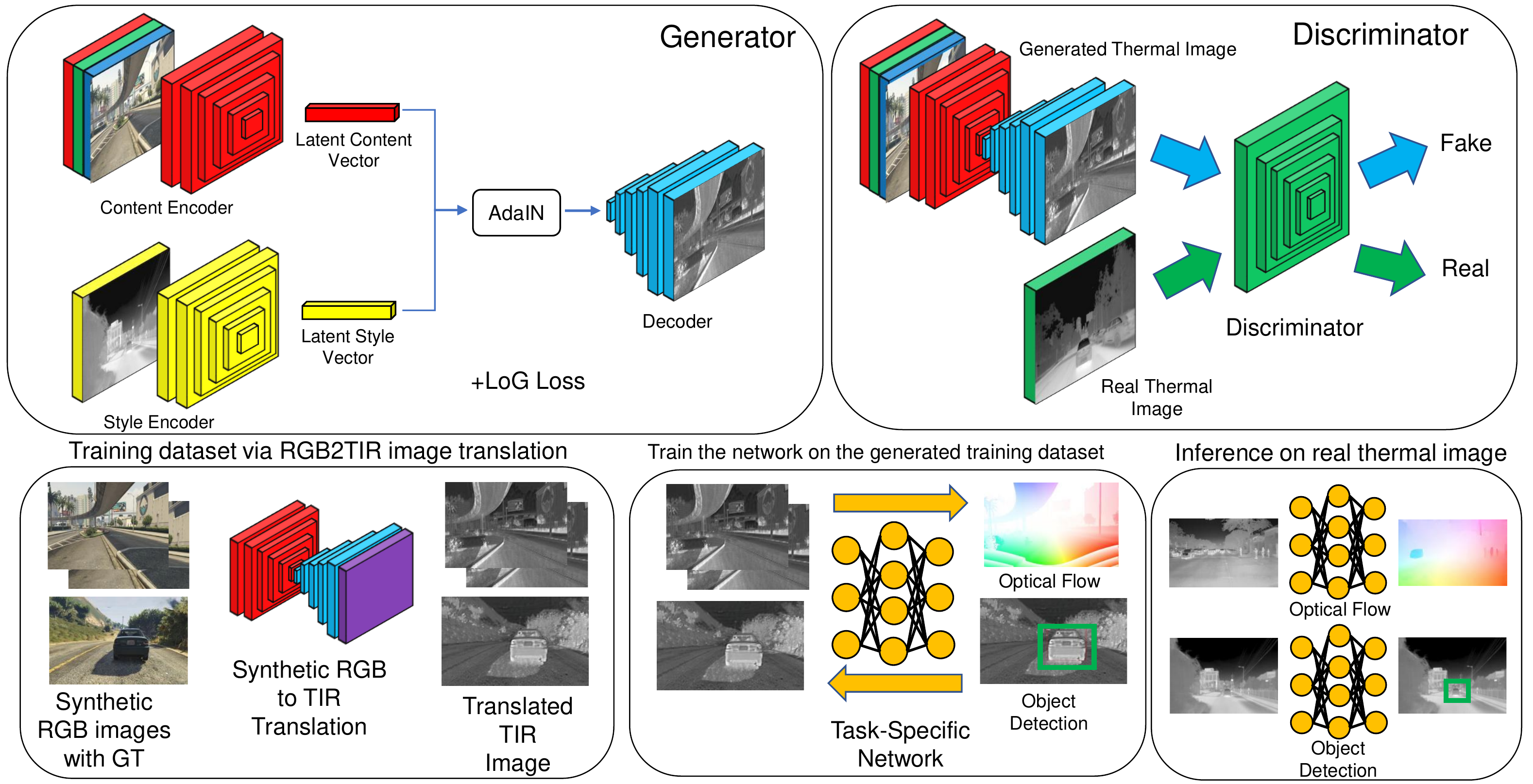}
  \end{center}
  \caption{RGB to TIR translation network and overall training pipeline. The encoders are divided into the content encoder (red) and a style encoder (yellow). Each encoder disentangles an image into each latent content and style vector. The decoder (blue) integrates the content and the style vector via adaptive instance normalization (AdaIN) \cite{adain} and generates the translated image. The role of the discriminator (green) is to classify the input image as either a real or a fake image. Using the translation model, synthetic RGB images with corresponding ground truths labels are translated into \ac{TIR} images, formulating \ac{TIR} image dataset with ground truths labels. Using the translated dataset, it can be leveraged to train any task-specific network in a supervised manner. Best viewed in color. }
  \label{network_architecture}
  \vspace{-8mm}
\end{figure*}

We adopted GAN-based multi-domain unpaired image to image translation architecture \cite{munit} as our baseline architecture. As illustrated in \figref{network_architecture}, the generator is composed of two encoders and a single decoder. Having contents and style vectors encoded from each encoder, the content encoder encapsulates the geometric attributes of an image such as edges and outlines of an image; the style encoder encapsulates the color and pixel intensity characteristics of \ac{TIR} images.




\subsubsection{Loss functions}
We employed adversarial loss \cite{goodfellow2020generative} for training GAN and style-augmented cyclic loss, $\mathit{L}_{cyc}$ \cite{munit}, to enforce additional cyclic consistency. We also used three types of reconstruction losses to enforce edge-guided and multi-domain translation. Although the translation is bidirectional, only the losses that are appropriate to RGB to \ac{TIR} translation will be expressed. 




\textbf{Image reconstruction Loss}: Concerns the ability of the translation network to reconstruct the original RGB image. 

\vspace{-2mm} \scriptsize
\begin{equation}
   \mathcal{L}_{recon}^{x_{RGB}} = \mathbb{E}[|| G_{RGB} (E_{RGB}^{c}(x_{RGB}),E_{RGB}^{s}(x_{RGB})) - x_{RGB}||_1 ]
\label{recon_loss_rgb}
\end{equation} \normalsize

\textbf{Content and Style Reconstruction Loss}: Related to the ability to extract valid latent content and style vectors from given images. As content vector from a RGB image is needed for generating the translated \ac{TIR} image, the content reconstruction loss computes the difference between the content vector of the original RGB image and that of the translated \ac{TIR} image. 

\vspace{-2mm} \small
\begin{equation}
   \mathcal{L}_{recon}^{c_{RGB}} = \mathbb{E}[|| E_{TIR}^{c} (G_{TIR}(c_{RGB},s_{TIR})) - c_{RGB}||_1 ]
\label{recon_content_loss_tir}
\end{equation} 
\normalsize

Similarly, the style reconstruction loss computes the difference between the style vector of the original \ac{TIR} image and that of the translated \ac{TIR} image. 

\vspace{-2mm} \small
\begin{equation}
   \mathcal{L}_{recon}^{s_{TIR}} = \mathbb{E}[|| E_{TIR}^{s} (G_{TIR} (c_{RGB},s_{TIR})) - s_{TIR}||_1 ]
\label{recon_style_loss_tir}
\end{equation} \normalsize


\textbf{Laplace of Gaussian (LoG) Loss}: Since unpaired RGB to \ac{TIR} image translation methods \cite{cyclegan,unit, munit} were severely under-constrained, we imposed additional constraints that can enforce the texture and edge consistency in the translated image. To achieve this, we utilized LoG loss \cite{dehazegan} between the original and the reconstructed image \gr{computes the distance between the extracted Laplacian features of the original and the reconstructed image.} It penalizes the network for generating images with second order gradient-based edge dissimilarity. The Laplacian features can be extracted by first applying a $3 \times 3$ Laplacian filter to each image channel \bu{ ($x_{TIR}^1$...$x_{TIR}^3$),}follow by global-average pooling the extracted features from each channel, like shown in \eqref{laplacian loss}. \bu{Unlike other edge-guided losses, LoG loss does not require any auxiliary edge extraction network for multi-task learning \cite{deep-edge-aware,luo2021edge}. 
}

\vspace{-2mm} \small
\begin{equation}
   \mathcal{L}_{Lap} = \mathbb{E}[|| L(x_{TIR}) - L(x_{TIR, recon})||_1 ]
\label{laplacian loss}
\end{equation}
\vspace{-5mm}
\begin{equation*}
    L(x_{TIR}) = \frac{1}{3}(L(x_{TIR}^1)+L(x_{TIR}^2)+L(x_{TIR}^3))
\end{equation*} \normalsize

\subsubsection{Overall training objective}

The overall training objectives of the network are shown in \eqref{total_loss}. All encoders, decoders, and discriminators are jointly trained and then optimized. The overall loss function is computed by a weighted sum of several different losses.

\vspace{-2mm} \scriptsize
\begin{eqnarray}
  \label{total_loss}
  \mathcal{L_G} &=& \mathcal{L_{GAN}}+\lambda_{x_{recon}}\mathcal{L}_{recon}^{x_{TIR}}+\lambda_{c_{RGB}} \mathcal{L}_{recon}^{c_{RGB}}+ \lambda{s_{TIR}} \mathcal{L}_{recon}^{s_{TIR}} \nonumber\\ 
  & &+\lambda_{Lap}\mathcal{L}_{Lap}+ \lambda_{cyc}\mathcal{L}_{cyc} 
\centering
\end{eqnarray}
\normalsize

For training the proposed image translation model, the loss weighting coefficients, $\lambda_{x_{recon}}, \lambda_{c_RGB}, \lambda_{s_TIR}, \lambda_{Lap}$, and $\lambda_{cyc}$ were set to 20, 10, 10, 20, and 5 respectively.

\subsection{Translation style selection}

\begin{figure}[t] 
  \begin{center}
    \includegraphics[width=\columnwidth]{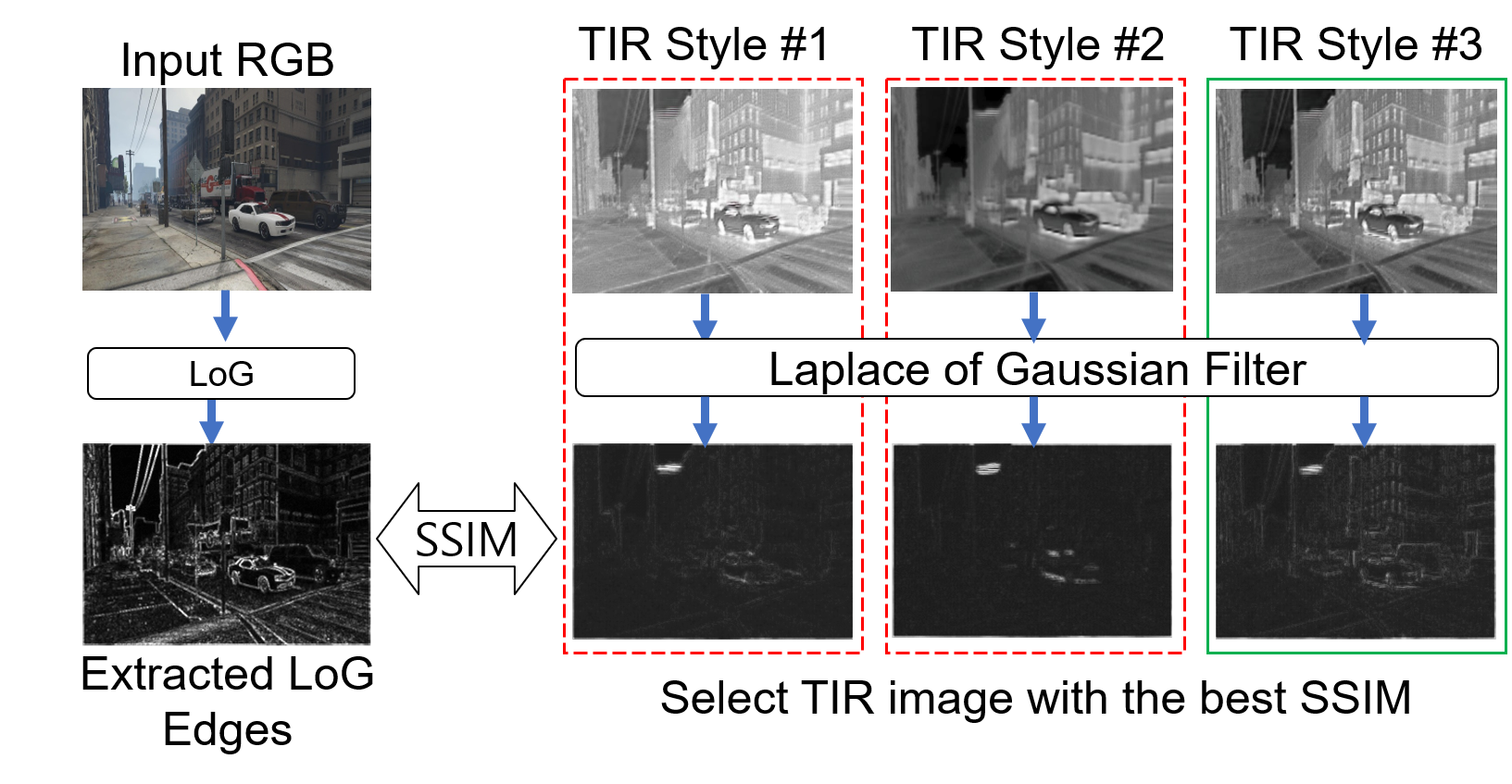}
  \end{center}
  \caption{Our proposed optimal style selection module. Edges of input RGB and several \ac{TIR} images are extracted from LoG filters, and SSIM between the RGB edge and each TIR edge are computed. TIR image with the best SSIM score is selected as the optimal style of the translation model.}
  \label{style_selection}
\end{figure}

A clear advantage of multidomain image to image translation (e.g. MUNIT) over bi-domain image translation (e.g. CycleGAN) is that the style of the translated image can be controlled with a given style code. By utilizing multi-domain translation, we can select style codes which has the least domain discrepancies to that of the input RGB image. By selecting the appropriate style from multiple sample style codes that were generated from various \ac{TIR} images, artifacts or erroneous patches can be minimized and characteristics of \ac{TIR} images can be accurately portrayed. 

As described in \figref{style_selection}, the style selection procedure is executed in the following way. First, using our proposed multi-domain RGB to \ac{TIR} translation model, we sample multiple translated \ac{TIR} images with different style codes. Second, we leveraged a $3\times3$ LoG filter to extract the edges from both the input synthetic RGB image and the translated \ac{TIR} images with different style codes.

Inspired by the works of \citeauthor{pearlgan} \cite{pearlgan}, we computed \ac{SSIM} between the extracted LoG edges of the input RGB image and each extracted LoG edge corresponding to the translated \ac{TIR} image with different style codes. We conjecture that higher \ac{SSIM} would refer to lesser number of artifacts created in the image but at the same time accurately portrayed the characteristics of \ac{TIR} image in the translated image well. 

For optical flow estimation, not only accurate \ac{TIR} characteristics need to be portrayed, especially for dynamic objects that correspond to ground truths flow, but the style of the translated images between the input image pair also needs to be consistent. We can achieve the input translated image pair consistency by translating the image pair with the same style codes.


\section{Experimental Setup}

\subsection{Synthetic RGB to \ac{TIR} image translation}
\subsubsection{Training dataset}

We leveraged several public benchmark datasets to train our proposed network in an unpaired manner. For RGB images, we utilized VIPER \cite{viper}. For \ac{TIR} images, we combined FLIR-ADAS \cite{flir-adas} and STheReO \cite{ssyun-2022-iros}. Overall, 13,554 RGB and 21,424 \ac{TIR} images were used. Details on the network architecture and parameters are available on our project page. 


\subsubsection{Training details}

We trained our proposed method against two bi-domain-based methods (CycleGAN \cite{cyclegan} and UNIT \cite{unit}) and a single multi-domain-based method, MUNIT \cite{munit}. All models were trained using only synthetic RGB images and real \ac{TIR} images. 

We used input image with resolution of $640 \times 400$ for both RGB and \ac{TIR} images 
For the network hyperparameters, we used Adam optimizer with a learning rate of 0.0001, weight decay of 0.5, 0.5 and 0.99 as $\beta_1$ and $\beta_2$ respectively. The network was trained for batch size of 1 for 60,000 iterations. 
\subsubsection{Evaluation criteria}

We evaluated our translation model using \ac{APCE} \cite{pearlgan}, which computes the average precision of extracted canny edges between RGB and the translated \ac{TIR} images at different thresholds. 

\subsection{Deep supervised \ac{TIR} optical flow estimation}

\subsubsection{\ac{TIR} image-based optical flow dataset}

To train the deep \ac{TIR} image-based optical flow estimation model, we first employed our proposed synthetic RGB to \ac{TIR} image translation model to the VIPER dataset. Overall, this yielded 13,356 and 4,954 optical flow image pairs and their corresponding flow ground truths labels for each training and validation data.

\subsubsection{Training details}

We employed recent state of the art deep optical flow estimation architectures, namely RAFT \cite{teed2020raft}, GMA-RAFT\cite{gma}, and GMFlow\cite{xu2022gmflow}. For training deep \ac{TIR} optical flow estimation models, we followed the practices of \cite{thermalraft} where we first train the model on synthetic RGB images for several epochs. Afterwards, we finetuned the trained model on our \ac{TIR} optical flow dataset. We employed random horizontal flip for data augmentations. As for the training hyperparameters, all three optical flow models were trained for 100,000 iterations at a batch size of 12 and were optimized using Adam optimizer with $\beta_1$ and $\beta_2$ as 0.5 and 0.99. Each RAFT, GMA-RAFT, and GMFlow leveraged learning rates of 0.0004, 0.00002, and 0.00002 with weight decay of 1e-4, 5e-5, and 5e-5 respectively.  

\subsubsection{Evaluation Criteria}

To quantitatively evaluate the performance of our \ac{TIR} optical flow estimation model, we translated validation optical flow dataset of VIPER into \ac{TIR} images. We evaluated our \ac{TIR} optical flow estimation model and preinitialized deep optical flow model on the TIR VIPER validation set using average end-point error (EPE), In addition, we use flow visualizations to qualitative evaluate the optical flow models on real \ac{TIR} images from several benchmarking \ac{TIR}-image datasets. 


\begin{figure*}[ht] 
  \begin{center}
    \includegraphics[width=\textwidth]{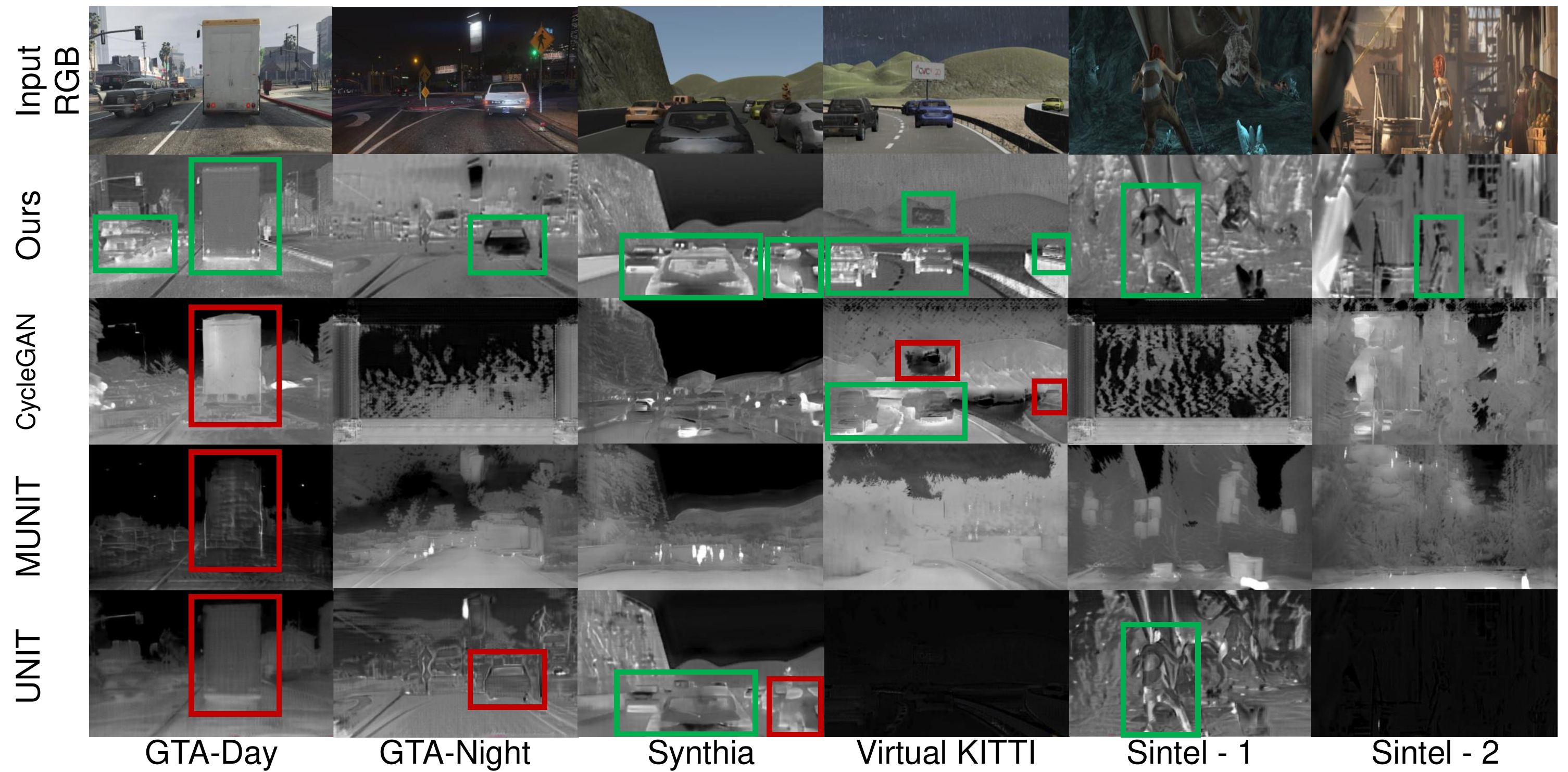}
  \end{center}
  \caption{Comparison of the baseline and the proposed method on synthetic RGB to \ac{TIR} image translation. Our proposed method has translated key details in the original image (green boxes) whereas the baseline methods fail to generate objects with sharp edges and realistic \ac{TIR} characteristics (red boxes). In addition, for several instances, the translation either completely fails or key objects within the image disappeared. Our translation method also generalizes well to unseen synthetic RGB datasets. Best viewed in color. }
  \label{gan_compare_synthetic}
\end{figure*}

\begin{figure*}[ht] 
  \begin{center}
    \includegraphics[width=\textwidth]{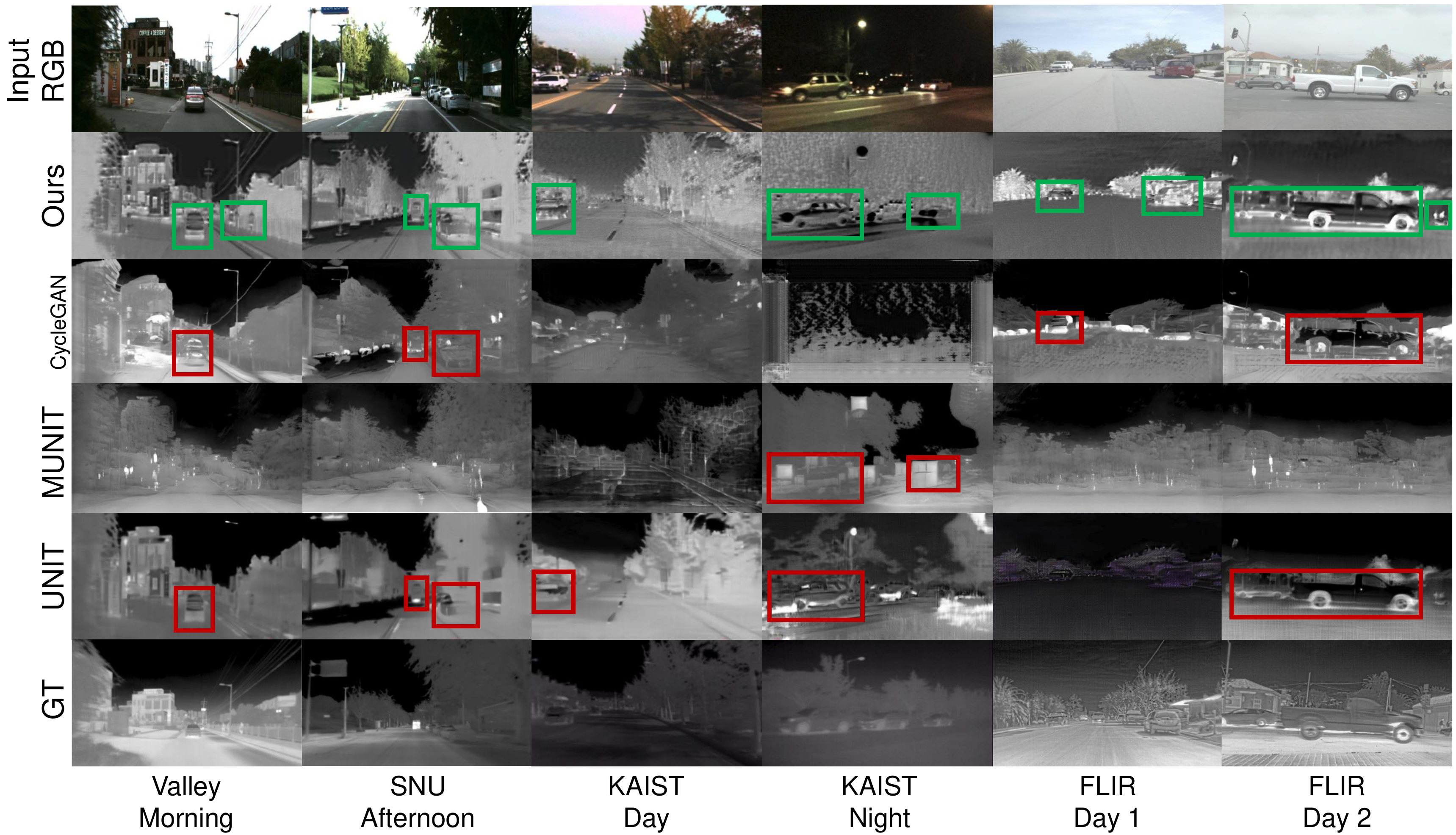}
  \end{center}
  \caption{Comparison of the baseline and the proposed method on real RGB to \ac{TIR} image translation. Even when trained on synthetic RGB images, our method generalizes well to real world RGB images regardless of the place and illumination condition, and characteristics of \ac{TIR} images are most accurately portrayed (green boxes). Same as before, artifacts and incorrectly translated objects are portrayed (red boxes); additionally, for some translation instances, key objects (green boxes) disappeared in the translated image. Best viewed in color. \bu{For better view of the images in the original size, please check out our project page.} }
  \label{gan_compare_real}
\end{figure*}

\section{RESULTS and Discussion}

\subsection{Edge-guided multi-domain RGB to TIR translation}

We evaluated our proposed methods and the baselines on RGB images from synthetic RGB image datasets (VIPER \cite{viper}, Synthia \cite{synthia}, Sintel \cite{sintel}, and Virtual Kitti \cite{kitti_flow}) and real world RGB image datasets (STheReO , KAIST \cite{hwang2015multispectral}, and FLIR-ADAS). 

The translation results for the generated \ac{TIR} images from synthetic and real RGB images are illustrated in \figref{gan_compare_synthetic} and \figref{gan_compare_real}. From a qualitative point of view, our proposed method outperformed other baselines in both synthetic RGB and real RGB image translations. 
The proposed method was able to maintain edge consistency as well as \ac{TIR} image characteristics, as indicated by the green boxes. In particular, unlike the poor translation results from other baseline methods, our method successfully maintained key details in Sintel-2, Virtual KITTI, and FLIR Day 1 images, and even small details presented in the FLIR Day 2 image were portrayed in the \ac{TIR} image generated using our proposed method. Similarly, the baseline methods either produced additional image artifacts or were unable to retain all the details presented in the original image, as indicated by the red boxes. However, our proposed method was able to translate all key visible details presented in the original RGB images. 

More importantly, one of the main bottlenecks of RGB to \ac{TIR} translation persisted in translating RGB image with poor illumination to \ac{TIR} images. In the baseline methods, poor \ac{TIR} image translation results were demonstrated for night time images (\figref{gan_compare_real}). However, \ac{TIR} image using our method has portrayed valid key details such as cars, showing better robustness to translation of night time images than other baseline methods. 



The qualitative evaluation of our proposed translation method can be supported by the quantitative \ac{APCE} evaluation. \tabref{edge_results} represents the average \ac{APCE} and per image \ac{APCE} evaluation for all models in both synthetic and real RGB image translation setting. According to the \ac{APCE} evaluation, our proposed model outperforms all other baselines in terms of average \ac{APCE} in both synthetic RGB and real RGB image translations; the same can be applied for per-image \ac{APCE} except for image translation on Virtual KITTI. Therefore, both qualitatively and quantitatively, our proposed translation method outperformed other existing method for translating from RGB to \ac{TIR} images.





\begin{figure}[] 
  \begin{center}
    \includegraphics[width=0.9\columnwidth]{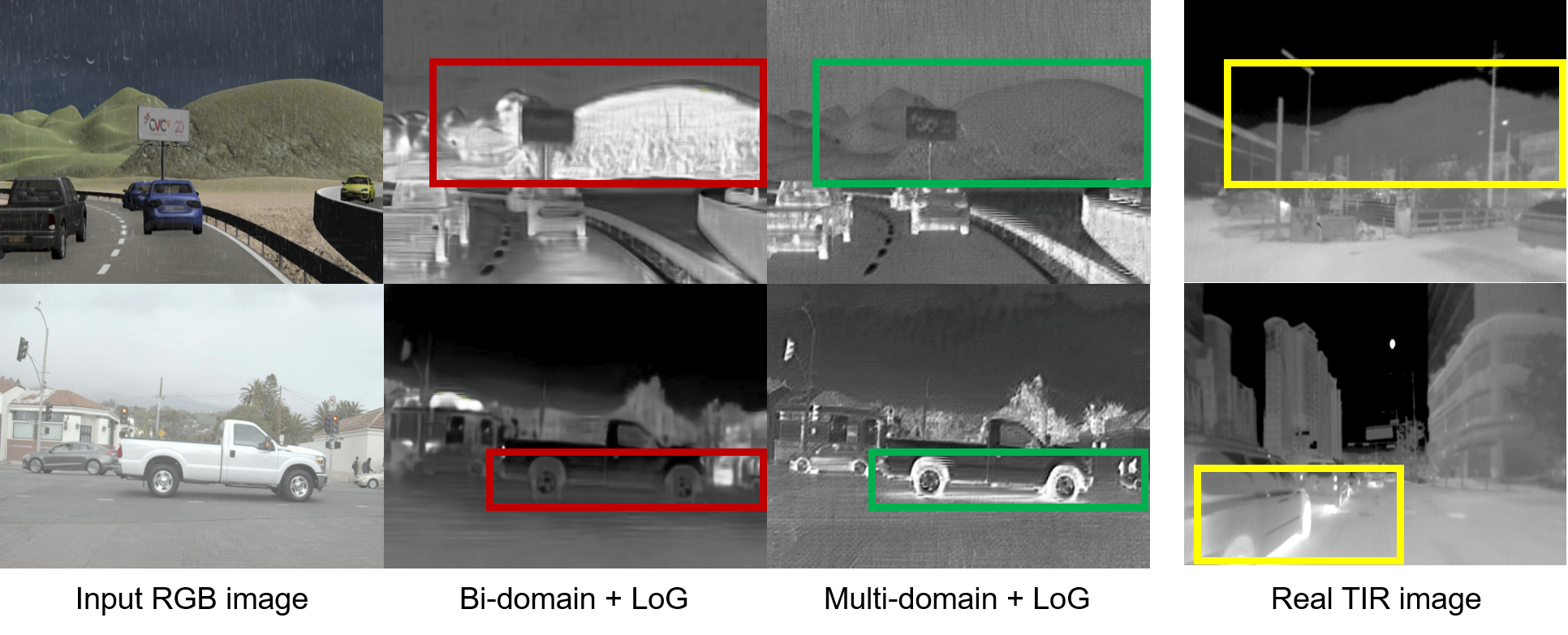}
  \end{center}
  \caption{Comparison of bi-domain model with LoG vs multi-domain with LoG. Realistic characteristics of \ac{TIR} images (yellow boxes) are well portrayed in the multi-domain methods (green boxes) but not in bi-domain-based methods (red boxes).} 
  \label{log_ablation}
  \vspace{-9mm}
\end{figure}

We performed further ablation study to examine the valid use of multi-domain model over bi-domain model in RGB to \ac{TIR} image translation. For evaluation, we compared our proposed method with a bi-domain model with LoG loss, as presented in \figref{log_ablation}.

\begin{table}[h]
\centering
\tiny 
\caption{APCE comparison for various methods on both real and synthetic images}
\label{edge_results}
\resizebox{0.8\columnwidth}{!}{%
\begin{tabular}{|ccccc|}
\hline
\multicolumn{1}{|c|}{Models}  & \multicolumn{1}{c|}{Proposed}        & \multicolumn{1}{c|}{CycleGAN}        & \multicolumn{1}{c|}{MUNIT}  & UNIT   \\ \hline
\multicolumn{5}{|c|}{Synthetic Images}                                                                                                             \\ \hline
\multicolumn{1}{|c|}{GTA-Day}       & \multicolumn{1}{c|}{\textbf{0.0793}} & \multicolumn{1}{c|}{0.0495}          & \multicolumn{1}{c|}{0.0175} & 0.0322 \\ \hline
\multicolumn{1}{|c|}{GTA-Night}       & \multicolumn{1}{c|}{\textbf{0.1264}} & \multicolumn{1}{c|}{0.0351}          & \multicolumn{1}{c|}{0.0262} & 0.0304 \\ \hline
\multicolumn{1}{|c|}{Synthia}       & \multicolumn{1}{c|}{\textbf{0.0967}} & \multicolumn{1}{c|}{0.0753}          & \multicolumn{1}{c|}{0.0289} & 0.0404 \\ \hline
\multicolumn{1}{|c|}{Virtual Kitti}       & \multicolumn{1}{c|}{0.1221}          & \multicolumn{1}{c|}{\textbf{0.1516}} & \multicolumn{1}{c|}{0.0165} & 0.0138 \\ \hline
\multicolumn{1}{|c|}{Sintel-1}       & \multicolumn{1}{c|}{\textbf{0.1301}} & \multicolumn{1}{c|}{0.0606}          & \multicolumn{1}{c|}{0.0226} & 0.0067 \\ \hline
\multicolumn{1}{|c|}{Sintel-2}       & \multicolumn{1}{c|}{\textbf{0.1246}} & \multicolumn{1}{c|}{0.0398}          & \multicolumn{1}{c|}{0.0118} & 0.0584 \\ \hline
\multicolumn{1}{|c|}{Average} & \multicolumn{1}{c|}{\textbf{0.11}}   & \multicolumn{1}{c|}{0.07}            & \multicolumn{1}{c|}{0.02}   & 0.03   \\ \hline
\multicolumn{5}{|c|}{Real Images}                                                                                                                  \\ \hline
\multicolumn{1}{|c|}{Valley-Morning}       & \multicolumn{1}{c|}{\textbf{0.0936}} & \multicolumn{1}{c|}{0.045}           & \multicolumn{1}{c|}{0.0232} & 0.0207 \\ \hline
\multicolumn{1}{|c|}{SNU Afternoon}       & \multicolumn{1}{c|}{\textbf{0.1502}} & \multicolumn{1}{c|}{0.0354}          & \multicolumn{1}{c|}{0.0248} & 0.0514 \\ \hline
\multicolumn{1}{|c|}{KAIST Day}       & \multicolumn{1}{c|}{\textbf{0.1683}} & \multicolumn{1}{c|}{0.0658}          & \multicolumn{1}{c|}{0.032}  & 0.0389 \\ \hline
\multicolumn{1}{|c|}{KAIST Night}      & \multicolumn{1}{c|}{\textbf{0.1432}} & \multicolumn{1}{c|}{0.0456}          & \multicolumn{1}{c|}{0.0399} & 0.046  \\ \hline
\multicolumn{1}{|c|}{FLIR Day-1}      & \multicolumn{1}{c|}{\textbf{0.0544}} & \multicolumn{1}{c|}{0.0395}          & \multicolumn{1}{c|}{0.0204} & 0.0281 \\ \hline
\multicolumn{1}{|c|}{FLIR Day-2}      & \multicolumn{1}{c|}{\textbf{0.0943}} & \multicolumn{1}{c|}{0.0486}          & \multicolumn{1}{c|}{0.0261} & 0.0451 \\ \hline
\multicolumn{1}{|c|}{Average} & \multicolumn{1}{c|}{\textbf{0.12}}   & \multicolumn{1}{c|}{0.05}            & \multicolumn{1}{c|}{0.03}   & 0.04   \\ \hline
\end{tabular}%
\normalsize
}
\end{table}

By using edge-guided training on the bi-domain model, valid key details were also depicted in the translated \ac{TIR} image. However, realistic characteristics of thermal images were hardly represented in the translated image. For example, in the bottom image, the \ac{TIR} image translated using the bi-domain method did not portray any \ac{TIR} residual heat marks  under the car; in contrast, using our proposed method based on multi-domain translation, characteristics of \ac{TIR} images were depicted in the translated image, indicated by the red boxes. Therefore, even with LoG loss being used to enforce edge in the image, bi-domain-based translation method hardly portrayed the correct characteristics of real \ac{TIR} images in the translated images. 

\subsection{Deep \ac{TIR} optical flow estimation}
\begin{figure}[] 
  \begin{center}
    \includegraphics[width=\linewidth]{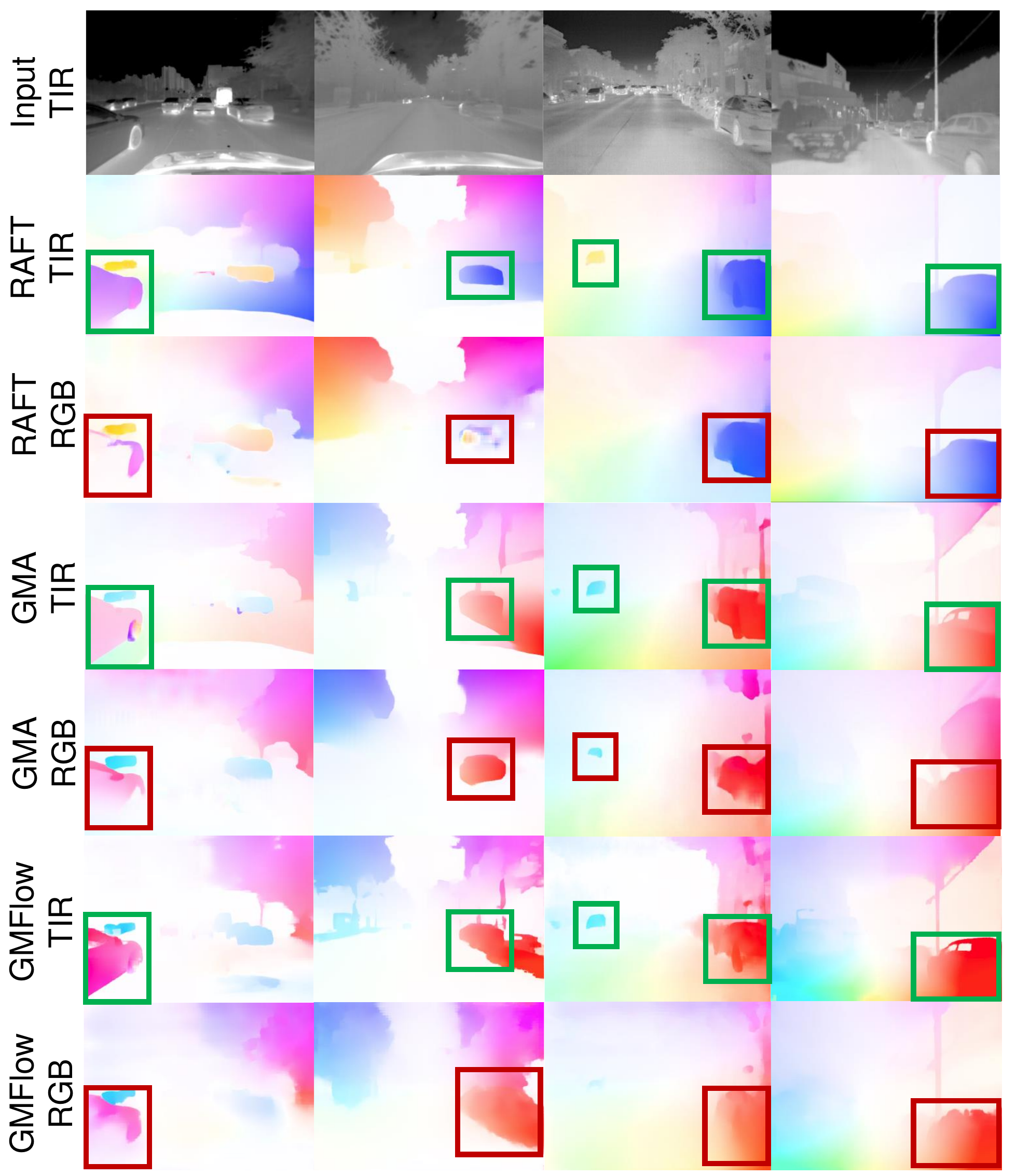}
  \end{center}
  \caption{Flow visualization comparison of various optical flow methods on real \ac{TIR} images. \ac{TIR} optical flow estimated by our trained models yielded sharper and more completed flow estimation, and they also yielded flow estimation from smaller objects (green boxes). In contrast, estimated flow from RGB-preinitialized methods resulted in occluded flow with ambiguous boundaries (red boxes). }
  \label{of_compare}
\end{figure}

The results to the quantitative and the qualitative evaluations are listed in \tabref{of-epe-eval} and \figref{of_compare}. From a qualitative evaluation on real \ac{TIR} images, our \ac{TIR} optical flow estimation models has shown better flow estimation performance than the baseline flow estimation model. Indicated by the green boxes, our \ac{TIR} optical flow model not only yielded sharper and clearer flow estimation on dynamic objects, but it also estimated flow from even smaller objects, which were not present in the preinitialized methods.

We observed ameliorated optical flow estimation performances in all three models that were finetuned on our \ac{TIR}-image dataset, outperforming the baseline preinitialized models. Out of all three models, GMA-RAFT displayed the lowest EPE, followed by RAFT and GMFlow. \gr{Despite the superior performance of GMFlow over RAFT in RGB image-based optical flow estimation, inferior performance was shown in \ac{TIR} images due to the insufficient number of \ac{TIR} image-based data to maximize the performance on transformer-based model.}

Since the characteristics of \ac{TIR} image greatly differs from that of RGB, flow estimation via conventionally used methods \cite{trigoni2020deeptio} has resulted in high EPE, yielding inaccurate flow estimation. On the contrary, estimating optical flow using our translated \ac{TIR} image dataset has yielded a valid performance improvement in real \ac{TIR} images. Particularly, the \ac{TIR} optical flow model was able to estimate flow from objects with ambiguous object boundary in the \ac{TIR} image. From this, we can deduce that our proposed translation method ameliorated \ac{TIR} optical flow estimation performance on real \ac{TIR} images when the model was only trained with generated \ac{TIR} data using our proposed method with RGB optical flow annotations. More importantly, this also demonstrated that our proposed translation model accurately portrayed \ac{TIR} characteristics in the generated \ac{TIR} images.

\begin{table}[!t]
\centering
\caption{ End point error evaluation on the our proposed and baseline methods}
\label{of-epe-eval}
\resizebox{0.7\columnwidth}{!}{%
\begin{tabular}{|cccc|}
\hline
\multicolumn{1}{|c|}{Model} & \multicolumn{1}{c|}{\begin{tabular}[c]{@{}c@{}}EPE\\ (RGB)\end{tabular}} & \multicolumn{1}{c|}{\begin{tabular}[c]{@{}c@{}}Training EPE\\ (Thermal)\end{tabular}} & \begin{tabular}[c]{@{}c@{}}EPE\\ (Thermal)\end{tabular} \\ \hline
RAFT                        & 43.762                                                                & 3.069                                                                                 & 16.18                                                   \\ \hline
GMFlow                      & 44.106                                                                   & 8.536                                                                                 & 24.568                                                  \\ \hline
GMA-RAFT                    & 41.740                                                                & \textbf{2.917}                                                                        & \textbf{15.733}                                      \\ \hline
\end{tabular}%
}
\end{table}

\begin{table}[h]
\centering
\caption{Object detection performance evaluation on FLIR Validation set }
\label{object_detection_map}
\resizebox{0.9\columnwidth}{!}{%
\begin{tabular}{|c|c|c|c|c|c|c|}
\hline
Methods  & mAP            & mAP\_50        & mAP\_75        & mAP\_S         & mAP\_M         & mAP\_L         \\ \hline
Proposed & \textbf{0.239} & \textbf{0.408} & \textbf{0.237} & 0.061          & \textbf{0.441} & \textbf{0.692} \\
CycleGAN & 0.007          & 0.01           & 0.01           & 0              & 0.005          & 0.009          \\
MUNIT    & 0.069          & 0.137          & 0.06           & 0.007          & 0.117          & 0.413          \\
UNIT     & 0.225          & 0.384          & 0.222          & \textbf{0.062} & 0.414          & 0.671          \\ \hline
\end{tabular}%
}
\end{table}

\subsection{Extension to object detection}

To further reveal the performance over semantic tasks, we examined the potential extension of our proposed method to object detection. \gr{For evaluation, we leveraged our proposed method and the baseline networks to generate \ac{TIR}-image based object detection datasets from a synthetic RGB object detection dataset \cite{johnson2017driving}, yielding four individual synthetic \ac{TIR} datasets. We then trained each dataset individually on VFNet \cite{zhang2021varifocalnet}}. We validated the trained networks on FLIR-ADAS \cite{flir-adas}. The detection performance of the trained models are displayed in  \tabref{object_detection_map}. 

According to the detection performance, our method  achieved the highest mean average precision (mAP) of 0.239 compared to other methods. Although UNIT has outperformed our proposed method in small object detection by 0.001, our proposed method has outperformed other baseline networks in other evaluation criteria. From these results, we confirm that our proposed method can not only be used for training tasks with challenging labels, but it also can be used to reduce effort for data annotation on other tasks like object detection.  




\section{CONCLUSION} \label{sec:conclusion}

Recent learning-based models in robotics and computer vision are becoming more data hungry. Especially, for those tasks with challenging labels such as optical flow, there are limitations to the number of annotated \ac{TIR} image datasets. This limitation can be addressed by using our edge-guided multidomain RGB to \ac{TIR} image translation network to generate annotated \ac{TIR} image dataset from synthetic RGB images. We confirmed the valid use of our proposed method through training deep \ac{TIR} optical flow and object detection by demonstrating performance improvements compared to other baselines. For further works, we plan to extend our approach to different tasks with challenging labels such as \ac{TIR} semantic segmentation and 3D object detection.






\small
\bibliography{ref}

\end{document}